# RCS-YOLO: A Fast and High-Accuracy Object Detector for Brain Tumor Detection

Ming Kang, Chee-Ming Ting(✉), Fung Fung Ting, and Raphaël C.-W. Phan

School of Information Technology, Monash University, Malaysia Campus, Subang Jaya, Malaysia
ting.cheeming@monash.edu

**Abstract.** With an excellent balance between speed and accuracy, cutting-edge YOLO frameworks have become one of the most efficient algorithms for object detection. However, the performance of using YOLO networks is scarcely investigated in brain tumor detection. We propose a novel YOLO architecture with Reparameterized Convolution based on channel Shuffle (RCS-YOLO). We present RCS and a One-Shot Aggregation of RCS (RCS-OSA), which link feature cascade and computation efficiency to extract richer information and reduce time consumption. Experimental results on the brain tumor dataset Br35H show that the proposed model surpasses YOLOv6, YOLOv7, and YOLOv8 in speed and accuracy. Notably, compared with YOLOv7, the precision of RCS-YOLO improves by 1%, and the inference speed by 60% at 114.8 images detected per second (FPS). Our proposed RCS-YOLO achieves state-of-the-art performance on the brain tumor detection task. The code is available at https://github.com/mkang315/RCS-YOLO.

## 1 Introduction

Automatic detection of brain tumors from Magnetic Resonance Imaging (MRI) is complex, tedious, and time-consuming because there are a lot of missed, misinterpreted, and misleading tumor-like lesions in the images of the brain tumors [8]. Most of the current work focuses on brain tumor classification and segmentation from MRI and detection tasks are less explored [1, 13, 22]. While existing studies showed that various Convolutional Neural Networks (CNNs) are efficient for brain tumor detection, the performance of using You Only Look Once (YOLO) networks is scarcely investigated [12, 20, 23–25, 27].

With the rapid development of CNNs, the accuracies of different visual tasks are constantly improved. However, the increasingly complex network architecture in CNN-based models, such as ResNet [6], DenseNet [9], Inception [28], etc. renders the inference speed slower. Though many advanced CNNs deliver higher accuracy, the complicated multi-branch designs (e.g., residual-addition in ResNet and branch-concatenation in Inception) make the models difficult to implement and customize, slowing down the inference and reducing memory utilization. The depth-wise separable convolutions used in MobileNets [7] also reduce the upper limit of the GPU inference speed. In addition, 3×3 regular convolution is highly optimized by some modern computing libraries. Consequently, VGG [26] is still heavily used for real-world applications in both research and industries.

RepVGG [2] is an extension of VGG via reparametrization to accelerate inference time. RepVGG uses a multi-branch topological architecture during the training phase, which is then reparameterized to a simplified single-branch architecture during the inference phase. In terms of the optimization strategy of network training, reparameterization was introduced in YOLOv6 [16], YOLOv7 [31], and YOLOv6 v3.0 [17]. YOLOv6 and YOLOv6 v3.0 employ reparameterization from RepVGG. RepConv, a RepVGG without an identity connection, is converted from RepVGG during inference time in YOLOv6, YOLOv6 v3.0, and YOLOv7 (named RepConvN in YOLOv7). Due to the removal of identity connections in RepConv, direct access to ResNet or the concatenation in DenseNet can provide more diversity of gradients for different feature maps. Grouped convolutions, which use a group of convolutions with multiple kernels per layer, like RepVGG, can also significantly reduce the computational complexity of the model, but there is no information communication between groups, which limits the ability of feature extraction of the convolution operator. In order to overcome the disadvantage of grouped convolutions, ShuffleNet V1 [34] and V2 [21] introduced the channel shuffle operation to facilitate information flows across different feature channels. In addition, when comparing Spatial Pyramid Pooling & Cross Stage Partial Network plus ConvBNSiLU (SPPCSPC) in YOLOv7 with Spatial Pyramid Pooling Fast (SPPF) in YOLOv5 [10] and YOLOv8 [11], it is found that more convolution layers in SPPCSPC architecture slow down the computation of the network. Nevertheless, SPP [4, 5] module achieves the fusion of local features and global features by max-pooling different convolution kernels' sizes in the neck networks.

Aiming at a faster and high accuracy object detector for medical images, we propose a new YOLO architecture called RCS-YOLO by leveraging on the RepVGG/RepConv. The contributions of this work are summarized as follows:

1) We first develop a RepVGG/RepConv ShuffleNet (RCS) by combining the RepVGG/RepConv with a ShuffleNet which benefits from reparameterization to provide more feature information in the training stage and reduce inference time. Then, we build an RCS-based One-Shot Aggregation (RCS-OSA) module which allows not only low-cost memory consumption but also semantic information extraction.
2) We design new backbone and neck networks of YOLO architecture by incorporating the developed RCS-OSA and RepVGG/RepConv with path aggregation to shorten the information path between feature prediction layers.

This leads to fast propagation of accurate localization information to feature hierarchy in both the backbone and neck networks.

3) We apply the proposed RCS-YOLO model for a challenging task of brain tumor detection. To our best knowledge, this is the first work to leverage on YOLO-based model for fast brain tumor detection. Evaluation on a publicly available brain tumor detection annotated dataset shows superior detection accuracy and speed compared to other state-of-the-art YOLO architectures.

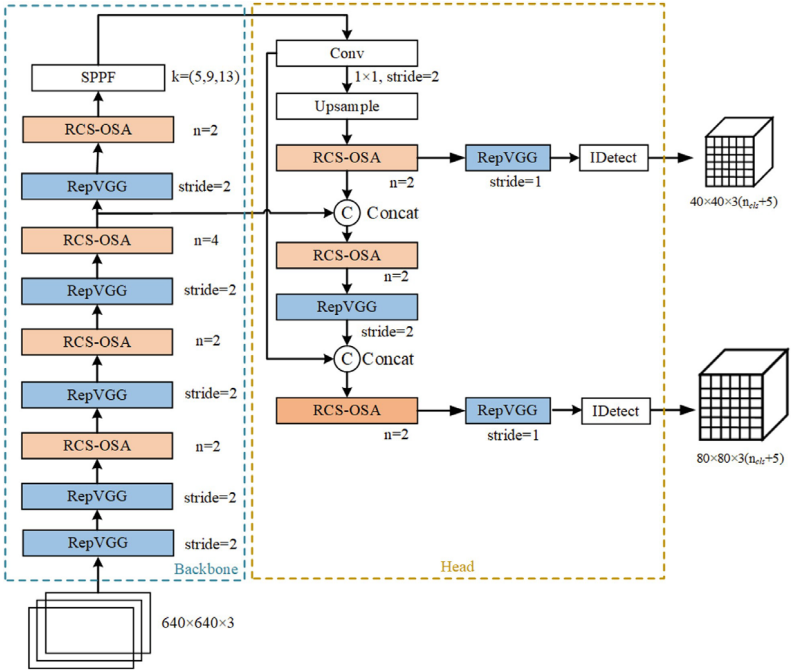

**Fig. 1.** Overview of RCS-YOLO. The architecture of RCS-YOLO is mainly comprised of the RCS-OSA (blue) and RepVGG (orange) modules. $n$ represents the number of stacked RCS modules. $n_{cls}$ represents the number of classes in detected objects. IDetect [29] from YOLOv7 denotes detection layers using 2D convolutional neural networks. (Color figure online)

## 2   Methods

The architecture of the proposed RCS-YOLO network is shown in Fig. 1. It incorporates a new module-RCS-OSA in the backbone and neck of the YOLO-based object detector.

### 2.1   RepVGG/RepConv ShuffleNet

Inspired by ShuffleNet, we design a structural reparameterized convolution based on channel shuffle. Figure 2 shows the structural schematic diagram of RCS.

Given that the feature dimensions of an input tensor are $C\times H\times W$, after the channel split operator, it is divided into two different channel-wise tensors with equal dimensions of $C\times H\times W$. For one of the tensors, we use the identity branch, 1×1 convolution, and 3×3 convolution to construct the training-time RCS. At the inference stage, the identity branch, 1×1 convolution, and 3×3 convolution are transformed to 3×3 RepConv by using structural reparameterization. The multi-branch topology architecture can learn abundant information about features during the training time, simplified single-branch architecture can save memory consumption during the inference time to achieve fast inference. After the multi-branch training of one of the tensors, it is concatenated to the other tensor in a channel-wise manner. The channel shuffle operator is also applied to enhance information fusion between two tensors so that the depth measurement between different channel features of the input can be realized with low computational complexity.

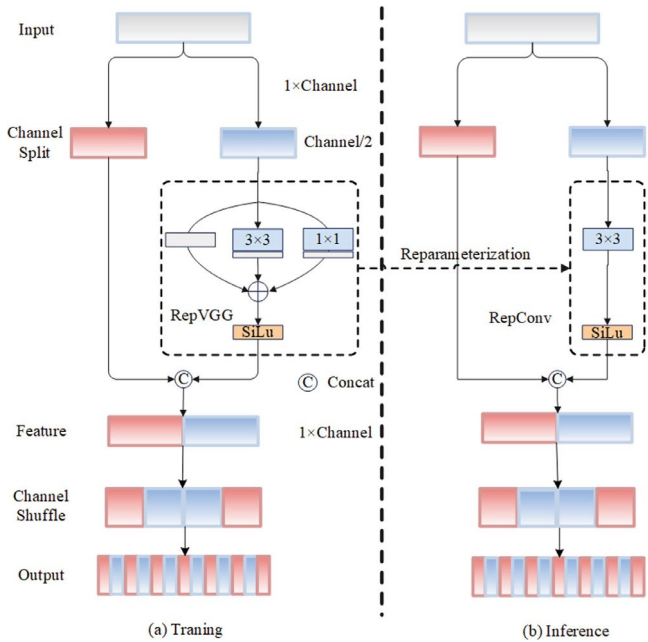

**Fig. 2.** The structure of RCS. (a) RepVGG at the training stage. (b) RepConv during model inference (or deployment). A rectangle with a black outer border represents the specific modular operation of the tensor; a rectangle with a gradient color represents the specific feature of the tensor, and the width of the rectangle represents the channel of the tensor.

When there is no channel shuffle, the output feature of each group only relates to the input feature within a group of grouped convolutions, and outputs from a certain group only relate to the input within the group. This blocks information flow between channel groups and weakens the ability of feature extraction. When channel shuffle is used, input and output features are fully related where one convolution group takes data from other groups, enabling more efficient feature information communication between different groups. The channel shuffle operates on stacked grouped convolutions and allows more informative feature representation. Moreover, assuming that the number of groups is $g$, for the same input feature, the computational complexity of channel shuffle is $\frac{1}{g}$ times that of a generic convolution.

Compared with the popular 3×3 convolution, during the inference stage, RCS uses the operators including channel split and channel shuffle to reduce the computational complexity by a factor of 2, while keeping the inter-channel information exchange. Moreover, using structural reparameterization enables deep representation learning from input features during the training stage, and reduction of inference-time memory consumption to achieve fast inference.

### 2.2 RCS-Based One-Shot Aggregation

The One-Shot Aggregation (OSA) module has been proposed to overcome the inefficiency of dense connections in DenseNet, by representing diversified features with multi-receptive fields and aggregating all features only once in the last feature maps. VoVNet V1 [14] and V2 [15] used the OSA module within its architecture to construct both lightweight and large-scale object detectors, which outperform the widely-used ResNet backbone with faster speed and better energy efficiency.

We develop an RCS-OSA module by incorporating RCS developed in Sect. 2.1 for OSA, as shown in Fig. 3. The RCS modules are stacked repeatedly to ensure the reuse of features and to enhance the information flow among different channels between features of adjacent layers. At different locations of the network, we set a different number of stacked modules. To reduce the level of network fragmentation, only three feature cascades are maintained on the one-shot aggregate path, which can mitigate the amount of network calculation burden and reduce the memory footprint. In terms of multi-scale feature fusion, inspired by the idea of Path Aggregation Network (PANet) [19], RCS-OSA + Upsampling and RCS-OSA + RepVGG/RepConv undersampling carry out the alignment of feature maps of different sizes to allow information exchange between the two prediction feature layers. This enables high-accuracy fast inference in object detection. Moreover, RCS-OSA maintains the same number of input channels and minimum output channels, thus reducing the memory access cost (MAC). For network building, we perpetuate max-pooling undersampling 32 times of YOLOv7 to construct a backbone network and adopt RepVGG/RepConv with a step of 2 to achieve undersampling. Due to the diversified feature representation of the RCS-OSA module and low-cost memory consumption, we use a different number of stacked RCS in RCS-OSA modules to achieve semantic information extraction during different stages of both backbone and neck networks.

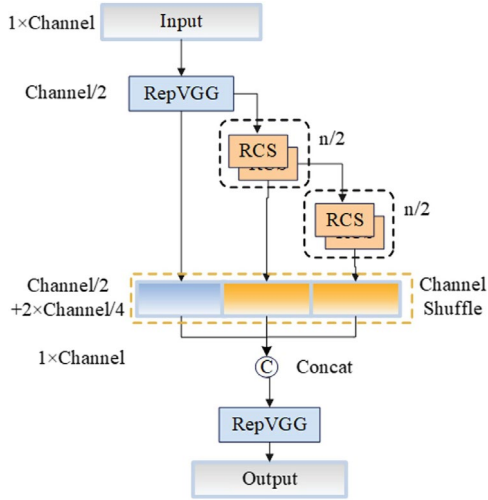

**Fig. 3.** The structure of RCS-OSA. $n$ represents the number of stacked RCS modules.

The common evaluation metric of computation efficiency (or time complexity) is floating-point operations (FLOPs). FLOPs are only the indirect indicator to measure the speed of inference. However, the object detector with a DenseNet backbone shows rather slow speed and low energy efficiency because the linearly increasing number of channels by dense connection leads to heavy MAC, which causes considerable computation overhead. Given input features of dimension $M \times M$, the convolution kernel of size $K \times K$, number of input channels $C_1$, and the number of output channels $C_2$, FLOPs and MAC can be calculated as:

$$FLOPs = M^2 K^2 C_1 C_2 \qquad (1)$$

$$MAC = M^2(C_1 + C_2) + K^2 C_1 C_2 \qquad (2)$$

Assuming $n$ to be 4, FLOPs of the proposed RCS-OSA and Efficient Layer Aggregation Networks (ELAN) [31, 33] are $20.25 C^2 M^2$ and $40 C^2 M^2$ respectively. Compared with ELAN, FLOPs of RCS-OSA are reduced by nearly 50%. The MAC of RCS-OSA (i.e., $6CM^2 + 20.25C^2$) is also reduced compared to that of ELAN (i.e., $17CM^2 + 40C^2$).

### 2.3 Detection Head

To further reduce inference time, we decrease the number of detection heads comprised of RepVGG and IDetect from 3 to 2. The YOLOv5, YOLOv6, YOLOv7, and YOLOv8 have three detection heads. However, we use only two feature layers for prediction, reducing the number of original nine anchors with different scales to four and using the K-means unsupervised clustering method to regenerate anchors with different scales. The corresponding scales are (87, 90), (127, 139), (154, 171), (191,240). This not only reduces the number of convolution layers and computational complexity of RCS-YOLO but also reduces the overall computational requirements of the network during the inference stage and the computational time of postprocessing non-maximum suppression.

## 3   Experiments and Results

### 3.1   Dataset Details

To evaluate the proposed RCS-YOLO model, we used the brain tumor detection 2020 dataset (Br35H) [3], with a total of 701 images in the 'train' and 'val' two folders, 500 images of which are the 'train' folder were selected as the training set, while the other 201 images in the 'val' folder as the testing set. For the input size of 640×640 image, the actual corresponding size is 44 32. The small object is defined as the object whose pixel size is less than 32×32 defined by the MS COCO dataset [18], so there are no small objects in the brain tumor medical image data sets, and the scale change of the target boxes is smooth, almost square. The label boxes of the brain images were normalized (See supplementary material Sect. 1).

### 3.2   Implementation Details

For model training and inference, we used Ubuntu 18.04 LTS, Intel® Xeon® Gold 5218 CPU processor, CUDA 12.0, and cuDNN 8.2. GPU is GeForce RTX 3090 with 24G memory size. The networking development framework is Pytorch 1.9.1. The Integrated Development Environment (IDE) is PyCharm. We uniformly set epoch 150, the batch size as 8, image size as 640× 640. Stochastic Gradient Descent (SGD) optimizer was used with an initial learning rate of 0.01 and weight decay of 0.0005.

### 3.3   Evaluation Metrics

In this paper, we choose precision, recall, $AP_{50}$, $AP_{50:95}$, FLOPs, and Frames Per Second (FPS) as comparative metrics of detection effect to determine the advantages and disadvantages of the model. Taking IoU = 0.5 as the standard, precision, and recall can be calculated by the following equations:

$$Precision = \frac{TP}{TP + FP} \quad (3)$$

$$Recall = \frac{TP}{(TP + PN)} \quad (4)$$

where $TP$ represents the number of positive samples correctly identified as positive samples, $FP$ represents the number of negative samples incorrectly identified as positive samples and $FN$ represents the number of positive samples incorrectly identified as negative samples. $AP_{50}$ is the area under the precision-recall (PR) curve formed by precision and recall. For $AP_{50:95}$, divide 10 IoU threshold of 0.5:0.05:0.95 to acquire the area under the PR curve, then average the results. FPS represents the number of images detected by the model per second.

### 3.4   Results

To highlight the accuracy and rapidity of the proposed model for the detection of brain tumor medical image data set, Table 1 shows the performance comparison between our proposed detector and other state-of-the-art object detectors. The time duration of FPS includes data preprocessing, forward model inference, and post-processing. The long border of the input images is set as 640 pixels. The short border adaptively scales without distortion, whilst keeping the grey filling with 32 times the pixels of the short border.

It can be seen that RCS-YOLO with the advantages of incorporating the RCS-OSA module performs well. Compared with YOLOv7, the FLOPs of the object detectors of this paper decrease by 8.8G, and the inference speed improves by 43.4 FPS. In terms of detection rate, precision improves by 0.024; $AP_{50}$ increases by 0.01; $AP_{50:95}$ by 0.006. Also, RCS-YOLO is faster and more accurate than YOLOv6-L v3.0 and YOLOv8l. Although the $AP_{50:95}$ of RCS-YOLO equals that of YOLOv8l, it doesn't obscure the essential advantage of RCS-YOLO. The results clearly show the superior performance and efficiency of our method, compared to the state-of-the-art for brain tumor detection. As shown in supplementary material Fig. 2, brain tumor regions are accurately detected from MRI by using the proposed method.

**Table 1.** Quantitative results of different methods. The best results are shown in bold.

| Model | Params | Precision | Recall | $AP_{50}$ | $AP_{50:95}$ | GFLOPs | FPS |
|---|---|---|---|---|---|---|---|
| YOLOv6-L [17] | 59.6M | 0.907 | 0.920 | 0.929 | 0.709 | 150.5 | 64.0 |
| YOLOv7 [31] | 36.9M | 0.912 | 0.925 | 0.936 | 0.723 | 103.3 | 71.4 |
| YOLOv8l [11] | 43.9M | 0.934 | 0.920 | 0.944 | **0.729** | 164.8 | 76.2 |
| **RCS-YOLO (Ours)** | 45.7M | **0.936** | **0.945** | **0.946** | **0.729** | **94.5** | **114.8** |

**Table 2.** Ablation study on proposed RCS-OSA module. The best results are shown in bold.

| Model | Params | Precision | Recall | $AP_{50}$ | $AP_{50:95}$ | GFLOPs | FPS |
|---|---|---|---|---|---|---|---|
| RepVGG-CSP (w/o RCS-OSA) | 22.6M | 0.926 | 0.930 | 0.933 | 0.689 | **43.3** | 6.1 |
| **RCS-YOLO (w/ RCS-OSA)** | 45.7M | **0.936** | **0.945** | **0.946** | **0.729** | 94.5 | **114.8** |

### 3.5   Ablation Study

We demonstrate the effectiveness of the proposed RCS-OSA module in YOLO-based object detectors. The results of RepVGG-CSP in Table 2, where RCS-OSA in the RCS-YOLO is replaced with the Cross Stage Partial Network (CSP-Net) [32] used in existing YOLOv4-CSP [30] architecture, are decreased than RCS-YOLO except GFLOPs. Because the parameters of RepVGG-CSP (22.2M) are less than half those of RCS-YOLO (45.7M), the computation amount (i.e., GFLOPs) of RepVGG-CSP is accordingly smaller than RCS-YOLO. Nevertheless, RCS-YOLO still performs better in actual inference speed measured by FPS.

## 4 Conclusion

We developed an RCS-YOLO network for fast and accurate medical object detection, by leveraging the reparameterized convolution operator RCS based on channel shuffle in the YOLO architecture. We designed an efficient one-shot aggregation module RCS-OSA based on RCS, which serves as a computational unit in the backbone and neck of a new YOLO network. Evaluation of the brain MRI dataset shows superior performance for brain tumor detection in terms of both speed and precision, as compared to YOLOv6, YOLOv7, and YOLOv8 models.